# Bridging AI and Clinical Practice: Integrating Automated Sleep Scoring Algorithm with Uncertainty-Guided Physician Review

Michal Bechny[1,2]

Giuliana Monachino[1,2]

Luigi Fiorillo[2]

Julia van der Meer[3]

Markus H. Schmidt[3,4]

Claudio L. A. Bassetti[3]

Athina Tzovara[1,5]

Francesca D. Faraci[2]

[1] Institute of Computer Science, University of Bern, Bern, Switzerland

[2] Institute of Digital Technologies for Personalized Healthcare (MeDiTech), University of Applied Sciences and Arts of Southern Switzerland, Lugano, Switzerland

[3] Department of Neurology, Inselspital, Bern University Hospital, University of Bern, Bern, Switzerland

[4] Ohio Sleep Medicine Institute, Dublin, United States;

[5] Center for Experimental Neurology, Department of Neurology, Inselspital, Bern University Hospital, University of Bern, Bern, Switzerland

Correspondence: Michal Bechny

Institute of Digital Technologies for Personalized Healthcare, East Campus USI-SUPSI, Via la Santa 1, CH-6962 Lugano-Viganello, Switzerland

Tel +41 (0)58 666 65 10

Email michal.bechny@supsi.ch

**Abstract**

**Purpose:**

This study aims to enhance the clinical use of automated sleep-scoring algorithms by incorporating an uncertainty estimation approach to efficiently assist clinicians in the manual review of predicted hypnograms, a necessity due to the notable inter-scorer variability inherent in polysomnography (PSG) databases. Our efforts target the extent of review required to achieve predefined agreement levels, examining both in-domain (ID) and out-of-domain (OOD) data, and considering subjects' diagnoses.

**Patients and methods:**

Total of 19578 PSGs from 13 open-access databases were used to train U-Sleep, a state-of-the-art sleep-scoring algorithm. We leveraged a comprehensive clinical database of additional 8832 PSGs, covering a full spectrum of ages and sleep-disorders, to refine the U-Sleep, and to evaluate different uncertainty-quantification approaches, including our novel confidence network. The ID data consisted of PSGs scored by over 50 physicians, and the two OOD sets comprised recordings each scored by a unique senior physician.

**Results:**

U-Sleep demonstrated robust performance, with Cohen's kappa (K) at 76.2% on ID and 73.8-78.8% on OOD data. The confidence network excelled at identifying uncertain predictions, achieving AUROC scores of 85.7% on ID and 82.5-85.6% on OOD data. Independently of sleep-disorder status, statistical evaluations revealed significant differences in confidence scores between aligning vs discording predictions, and significant correlations of confidence scores with classification performance metrics. To achieve κ ≥ 90% with physician intervention, examining less than 29.0% of uncertain epochs was required, substantially reducing physicians' workload, and facilitating near-perfect agreement.

**Conclusion:**

Inter-scorer variability limits generalizability of scoring algorithms to an accuracy of about 80%. Therefore, a thorough review of predicted hypnograms by physicians remains imperative, to align with the scoring taste of the responsible physician. Our work offers an approach to conduct this review efficiently and demonstrates validity across ID and OOD data, considering individuals' clinical status.

**Keywords:** automated sleep scoring, uncertainty quantification, explainable AI, polysomnography, sleep medicine

# Introduction

Sleep, often dubbed as the third pillar of health alongside diet and exercise, plays a critical role in our well-being. Polysomnography (PSG), a comprehensive sleep monitoring technique, captures detailed biosignals - primarily the electroencephalogram (EEG), the electrooculogram (EOG), and the electromyogram (EMG). Adhering to guidelines of American Academy of Sleep Medicine (AASM),[1] physicians score PSG recordings into specific sleep stages, on 30-second windows (*epochs*). Such structured scoring, called *hypnogram*, divides sleep into five distinct stages: W, REM, N1, N2, and N3, each representing a unique physiological state.[2] The proportions of sleep stages, as well as patterns in their transitions, are basic indicators of sleep health,[3,4] and also biomarkers of certain disorders.[5-7]

While manual scoring remains the gold standard, the procedure may be labor-intensive, often demanding up to 2 hours for a comprehensive evaluation of a single PSG recording.[8] Research into automatic sleep scoring, which aims to support the manual scoring of physicians by computational algorithms, dates back to the 1960s.[9] Recent advancements in Artificial Intelligence (AI) have significantly improved automatic scoring solutions, especially those based on Machine and Deep Learning (ML/DL) methodologies. Notably, the U-Sleep algorithm introduced in 2021,[10] stands at the forefront due to its balance between scoring performance and the diversity of its training data. It was optimized on a vast collection of 16 clinical databases with over 15000 subjects, and demonstrated epoch-wise performance rivaling human scorers, achieving an average F1-score of 79%. Furthermore, Fiorillo & Monachino et al (2023) conducted extensive tests of network's resilience and confirmed U-Sleep robustness by reporting on average an F1-score of 76.5% on 13 open-access databases.[11]

Supervised automated sleep scoring algorithms can reach considerable performance but are to-date not able to overcome an intrinsic problem. The different interpretations of AASM scoring standards by physicians result in an inter-scorer agreement of about 76%.[12-14] This human-based variability in the annotations introduces approximately 20% noise-level, technically limiting the performance of scoring algorithms optimized in a supervised way, as the ability of an AI algorithm can hardly be better than the quality of its training data. Consequently, despite the breadth of training databases available, the ceiling for ML/DL model generalizability is limited by this prevailing inter-scorer agreement. Therefore, despite the technological advancements AI has brought to sleep scoring, physicians - who are still irreplaceable and responsible for clinical decisions - must subject the predicted hypnograms to a thorough review and compare whether the algorithm-proposed predictions are consistent with their personal interpretation of patterns present in the original PSG biosignals. While some level of error in sleep-scoring models is deemed clinically acceptable,[15] the review process of predicted hypnograms can be time-consuming and costly. Specifically, if physicians lack prior insights into problematic segments of the biosignal, the review might be as resource intensive as conducting manual scoring without any algorithmic assistance.

Given the limits posed by inter-scorer variability, a subset of research has pivoted towards quantifying prediction uncertainty in order to elevate model performance by enabling review of the least confident predictions. Such semi-automated approaches combining predictions proposed by algorithms with physician's expertise represent a promising solution for integration of sleep scoring tools in clinical setting.[9] Van Gorp et al (2022) delved into the theoretical aspects of such (un)certainty.[16] Kang et al (2021) advanced this notion by proposing an uncertainty detection mechanism via Shannon's entropy of the softmax output of a statistical classifier.[17] By allowing physicians to correct uncertain predictions, they managed to substantially enhance the agreement (K-score) between classifier and physician's scoring taste. In the realm of DL-based algorithms, Fiorillo et al (2021) employed a query procedure targeting a predetermined percentage of the most uncertain predictions based on the maximum and variance of the softmax output.[18] Hong et al (2021) presented a novel method, Dropout-Correct-Rate, and showcased its potential to boost model performance with targeted human review.[19] Meanwhile, Phan et al (2022) utilized a transformer-based sleep scoring model and identified uncertain epochs through normalized entropy scores, demonstrating that a substantial fraction of misclassified predictions were within the most uncertain epochs.[20] Most recently, Rusanen et al (2023) evaluated several softmax-based measures of aSAGA, a convolutional neural classifier, and reported effective identification of predictions in the mismatch to the consensus-scoring of 5 scorers.[21]

The integration of sleep-scoring algorithms into clinical practice demands a deep understanding of the physician's real needs and expectations. However, these are seldom considered in existing work aiming to automate sleep scoring, which approaches this problem in isolation from the human experts. Our study builds upon the U-Sleep algorithm, a state-of-the-art DL-based sleep scoring model trained on a broad spectrum of open-access clinical databases. Considering the intrinsic limitations of sleep scoring, rather than just aiming to improve the model's epoch-wise performance, which might already be at its ceiling level due to the inter-scorer variability, our study seeks to integrate this established system in a manner that actively involves physicians.

By investigating various strategies for pinpointing the least confident predictions and streamlining their review, we aim to redefine the collaboration between sleep-scoring algorithms and clinicians. Utilizing clinically rich *Berner Sleep Data Base* (BSDB),[22] we systematically investigate (i) *the optimal strategies to gather uncertain sleep stage predictions for the physicians' review* and based on that we (ii) *quantify the volume of predictions that need to be reviewed (ie, physician's effort) to reach certain agreement benchmarks*.

The BSDB contains PSGs scored by more than 60 different scorers of different experience levels and covers subjects from a full spectrum of ages and sleep-disorders, enriching our evaluation framework. Utilizing the information about PSG-scorers, we assess the efficacy of the combined system integrating the sleep-scoring algorithm with uncertainty estimation, considering both *in-domain* (ID) and *out-of-domain* (OOD) test data, and accounting also for subjects' diagnoses.

Semi-automated approaches for sleep staging have been proposed and explored in various modalities and frameworks.[9,16-21] However, comprehensive testing addressing all possible limitations of these methods has been relatively sparse. To the best of our knowledge, our research is the first to conduct extensive tests addressing a broad range of challenges specific to semi-automated sleep scoring. This includes an in-depth examination of individual scoring tastes of single physicians, the impact of different sleep-disorder diagnoses on approach validity, the adopted metrics, as well as the dimensions and diversity of the datasets involved. Our approach advances the understanding of sleep scoring solutions, considering a wide array of factors that influence their effectiveness and practicality for clinical integration.

# Material and methods

## *Dataset*

For the evaluations, we exploited the *Berner Sleep Data Base (BSDB)* registry from our partner clinic, Inselspital, University Hospital Bern. More than 8000 PSGs have been collected from 2000 to 2021 on individuals covering the whole spectrum of age (0-91 years), sleep disorders, as well as healthy controls. The signals were recorded at 200 Hz and, across 20 years of data collection, scored manually by more than 10 senior and 50 assistant physicians according to the AASM rules. Secondary usage was ethically approved (KEK-Nr. 2020-01094). Most individuals underwent PSG due to the suspicion of a sleep disorder. Together 66 individuals represented healthy subjects that took part as controls in clinical trials. The BSDB provides various information on diagnoses based on individual tests (eg, actigraphy- or PSG-based), however, in our evaluations, we used only the clinically most relevant *conclusive diagnoses* made by physicians considering all test-based diagnoses, clinical anamnesis, and the context.

For the purpose of our research, we divided the BSDB into three parts: one in-domain (ID) subset - consisting of training, validation, and test data splits consisting of PSGs, each scored by one of >50 physicians - used for optimization and baseline evaluation of the model-based approaches adopted - and, utilizing the information about the scorers, we created two out-of-domain (OOD) held-out subsets, each containing PSGs scored by a unique senior physician not presented in ID data with potentially different "scoring taste" than the population of ID-included physicians. Hence, the evaluations done on OOD subsets represent a more robust generalizability assessment close to the scenario happening in clinics, where typically a single physician takes decisions (eg, about scoring, diagnosis). As one individual can have multiple PSGs recorded, all data splits were done per subject, assuring that the individual's data are present only in one subset. A summary of data splits with respect to the number of PSGs, physicians involved, age, and gender is provided in Table 1. In addition, Table 2 provides details on occurrence of different classes of sleep disorders among conclusive diagnoses of subjects.

## *U-Sleep: the sleep scoring algorithm*

The U-Sleep, introduced by Perslev et al (2021),[10] is a deep convolutional neural network for sleep stage classification inspired by the U-Net, an architecture originally used for image segmentation.[23] The U-Sleep takes as its input at least one pair of EEG-EOG channels (re)sampled at 128 Hz and outputs an array of softmax values quantifying the plausibility of each signal window (epoch) of a specified length, usually 30 seconds, to represent one of the 5 sleep stages. If more than one channel pair is used, the U-Sleep applies majority voting that averages the softmax outputs over all channel pairs used. The architecture of U-Sleep consists of an encoder-decoder part followed by the classifier layer. The encoder part compresses the input signal into a lower-resolution representation using subsequent convolutional operations followed by nonlinear activation functions. The decoder part then transforms this compressed representation back to the original time resolution using convolution operations and up-sampling. The final part of the architecture consists of an average-aggregation layer and the softmax layer that outputs predicted probabilities on the desired window size.

In-depth technical details on the U-Sleep architecture, implemented preprocessing, and training can be found in the original work,[10] that also reports the state-of-the-art performance on 16 databases of more than 15000 participants, achieving an average F1-score of 79%. The robustness of U-Sleep was confirmed even after its original implementation was corrected for a channel-derivation bug, achieving an average F1-score of 76.5%.[11]

Part of our work replicated the training run on 13 open-access databases of the most recent implementation of U-Sleep.[11] Based on that, we exploit the rich BSDB and fine-tune (re-train) the U-Sleep using training and validation ID-splits as described in Table 1. Finally, we use such fine-tuned U-Sleep as a basis for the selection of the most suitable approach of uncertainty estimation to enable an efficient review of predicted hypnograms by physicians. The generalizability of both sleep scoring as well as (predictive) uncertainty quantification approaches is in-depth examined on the ID-test and the two OOD parts of BSDB.

## *Estimation of predictive uncertainty*

In advancing sleep scoring algorithms for clinical practice, one crucial component is the quantification of predictive uncertainty, which encompasses both *epistemic* and *aleatoric* aspects. Epistemic uncertainty, in sleep-scoring context, arises from the variability in how physicians interpret AASM guidelines, leading to about 20% noise in sleep-stage labels due to approximately 80% inter-scorer agreement. On the other hand, aleatoric uncertainty, inherent in the variability of sleep patterns themselves, represents the natural randomness that cannot be mitigated.

In this section, we elaborate on our approach with the U-Sleep classifier. First, we detail measures of predictive uncertainty based on the classifier's softmax output. Subsequently, we describe our efforts in adapting an auxiliary confidence network, specifically designed for sleep-related time-series representations derived from the U-Sleep, to estimate predictions' confidence. The terms uncertainty and confidence can be understood as complementary and will be used according to the appropriateness of the context. The integration of uncertainty quantification is pivotal not only in elevating the trustworthiness of the automated sleep-scoring solutions but also in enabling physicians to efficiently review and verify algorithm-proposed predictions.

## Softmax-based measures

The confidence level of classifier's predictions can be gauged from its softmax output, which can be graphically represented as so-called *hypnodensity*.[24] This output can be analyzed either visually, or, when uncertain epochs should be automatically gathered, by numerical assessment of the softmax values. At its simplest, the maximum value of the predicted softmax can be perceived as a representation of the epoch's likelihood of belonging to a specific class. The closer the max-softmax value is to 1, the higher the confidence, while lower values indicate uncertainty. There are a variety of measures, rooted in softmax outputs, that can be employed to discern these uncertainties. For instance, several works employed entropy-based measures because as entropy rises, the distribution of softmax values becomes more uniform.[17,20,21]

Regardless of the chosen measure, uncertain predictions from each predicted hypnogram can be highlighted in two ways: *(i) by showcasing a fixed percentage of the most uncertain epochs*, or *(ii) by indicating epochs that surpass a specific value threshold.* The latter is more advocated as it may consider sampling distribution of classification accuracy. Moreover, the fixed-percentage approach has greater potential to introduce undesired results (false positives/negatives) if the predetermined percentage does not coincide with the actual amounts of misclassified epochs. In our research, we sought methods that adeptly identify uncertain predictions for subsequent review by clinical experts. A comprehensive mathematical detailing of all measures employed in our work is provided in Table 3, whereas the comparison in terms of their ability to discern predictions discordant with human scoring is presented in Results.

## Uncertainty quantification using an auxiliary confidence network

Neural networks, while powerful, often exhibit overconfidence, manifested as a disparity between the predicted softmax value and the actual probability of an observation belonging to a specific class.[25] This may bring possible limitations for the use of softmax-based hypnodensity to gather uncertain predictions accurately. To counteract the overconfidence issue, Corbiere et al (2021) proposed an auxiliary confidence network, which aims to estimate the *True Class Probability* (TCP) score, designed to work in tandem with an already-trained classifier network.[26] Intriguingly, the TCP is defined as the value of the predicted softmax that aligns with the true label, meaning, for misclassified predictions, it diverges from the softmax maximum value. Upon the completion of classifier training, the TCP for each observation is extracted from training and validation data and serves as a target for the confidence network. This positions the training of the confidence network as a regression problem, where the objective is to predict the TCP - a single float value within the (0, 1) range - for each observation. In the original work, the confidence network was applied to image data, supplementing a convolutional network classifier, which involved reusing the classifier's architecture and its pre-trained weights, adding additional layers to the classifier to facilitate the prediction of the numerical TCP outcome, and then fine-tuning the modified architecture to optimize the TCP predictions.[26]

Our contribution extends this idea specifically to PSG time-series data. Leveraging the U-Sleep output, we designed a lightweight sequence-to-sequence long-short-term-memory (LSTM) confidence network.[27] For each EEG-EOG input channel-pair of U-Sleep, our confidence network is fed by representations extracted from U-Sleep layers, including the 5-dimensional softmax output, the binary code of the same dimensionality as softmax indicating the predicted class, and the 5-dimensional hidden features extracted from layer preceding the softmax. The adoption of an bidirectional-LSTM-based architecture was driven by our beliefs that the uncertainty in predicting sleep stages is intrinsically tied to sequential information - namely, the representations preceding and succeeding a given epoch. Recognizing the functional dependencies in the softmax output (that sums up to 1), we applied to it the additive log-odds ratio (ALR) transformation, which reduces the dimension by one (ie, to 4) and decreases the co-linearity.[28] Building on the premise that combined data offers a richer perspective for identifying the most uncertain predictions, we fed the confidence network with all such extracted features simultaneously. The final architecture of our confidence network had 35628 parameters and consisted of three main parts: an input layer with batch normalization; 4 hidden layers (LSTM of 50 neurons, bidirectional-LSTM of 30 neurons enabling information flow from the past as well as future states, two LSTMs of 10 and 5 neurons) returning sequences, with tanh activation function and 25% drop-out; and a final layer with an output LSTM neuron with the custom activation function, $(\tanh(x)+1)/2$, returning a sequence in desired (0, 1) range, corresponding to the predicted sequence of TCP scores for to each epoch. These are then, consistent with U-Sleep's majority voting mechanism, averaged across all EEG-EOG input channel-pairs used. Confidence scores predicted by the network extended the set of softmax-based uncertainty measures, and our evaluations focused on in-depth comparison of their utilities for identifying discordant predictions from human scoring.

## *Utilizing uncertainty estimates for an efficient review of predicted hypnograms*

Our analysis, tailored towards the efficient use of uncertainty estimates for the review of predicted hypnograms, was guided by a three-tiered evaluation approach: (i) selection of the best-suited uncertainty measure; (ii) statistical evaluations of its discriminative power; and (iii) the impact-evaluation when physicians rescore the most uncertain predictions gathered. While the first two aspects focus on the technical aspects, the conclusive part evaluates the practical implications, comparing the physician's effort - quantified as the amounts of epochs reviewed - in relation to the boost of the agreement between their scoring taste and partially reviewed predictions of the scoring algorithm.

### Best-suited uncertainty measures

Initially, to pinpoint the most suitable uncertainty measure, we treated identifying epochs diverging from human scoring as a binary classification task. The diverging epochs from human scoring, ie, the U-Sleep-misclassified predictions, were considered as a positive class. Using this setup, we selected the most apt measure based on their Receiver Operating Characteristic (ROC) and Precision-Recall (PR) curve performances. The choice of ROC and PR curves stems from their ability in handling class imbalances and effectively comparing the true-positive against false-positive rates.

## Statistical tests to assess the discriminative power of the superior uncertainty metric

Upon identifying the superior metric, we further sought to statistically assess its efficacy in two distinct manners. Firstly, we proposed the null hypothesis $H_{01}$: *"There is no significant difference between the on-subject mean-aggregated uncertainty scores of epochs congruent with human scoring and those diverging from it."* In other words, this would imply that the uncertainty in correctly scored epochs would be the same as for the misclassified ones. With $H_{01}$ we aimed to test whether predictions in line with human scoring systematically differed from those diverging in terms of their uncertainty score, effectively probing the metric's ability to distinguish between correctly versus incorrectly classified epochs.

Further, the null hypothesis $H_{02}$ postulated: *"There is no significant correlation between the mean-aggregated on-subject uncertainty scores and the on-subject classification performance metrics."* In other words, that would imply that, eg, classification accuracy is not associated with uncertainty levels. The $H_{02}$ aimed to assess the relationship between the uncertainty attached to predictions and the classification performance on a per-subject basis.

Both assessments were conducted separately for ID and OOD data, with consideration of sleep-disorder status of individuals. Given the skewed non-normal nature of the uncertainty measures with bounded value ranges, the non-parametric bootstrap was employed to calculate confidence intervals (CI) in order to assess both hypotheses.[29]

## Impact-evaluation of physician intervention on uncertain epochs

The culmination of our analysis revolved around varying the threshold employed to discern the uncertain epochs for the superior uncertainty metric identified. Under each threshold specification from a predefined grid, a physician review was enacted, with discordant predictions being rectified and agreeing epochs being kept. Subsequently, the classification metrics were recalculated to encapsulate this simulated physician's intervention. While the relation between increased reviewed epochs and monotonic performance improvement is evident, our objective was to quantify the rescoring effort required to meet distinct performance benchmarks. This examination was undertaken across both ID and OOD test data splits, fortifying the robustness of our conclusions. Further, in order to make fair comparisons with existing research, we enumerated the performance improvements across diverse metrics: accuracy (Acc), weighted F1-score ($F1_w$), and Cohen's kappa (K).

# Results

In this section we provide the main findings with respect to the algorithmic methods exploited and developed (U-Sleep algorithm along with the auxiliary confidence neural network), and their validation on individual data domains, as depicted within the workflow in Figure 1.

## *U-Sleep classification performance*

As a sleep scoring classifier, we employed U-Sleep and replicated the training experiment of its most recent implementation using 13 open-access databases of 19578 PSGs.[11] Next, the model was fine-tuned on the BSDB, leveraging the ID training and validation splits as elaborated in Table 1. The U-Sleep optimization based on minimization of the categorical cross-entropy loss converged after 539 training epochs. To ensure a comprehensive comparison with existing research, we enumerated three distinct classification performance metrics: Acc, $F1_w$, and K, computed in three different ways: epoch-wise (pertaining to all 30-second windows in the relevant data split), as well as subject-wise mean- and median- aggregated. Table 4 summarizes the performance across the ID and the two OOD test data. The results indicate that the epoch-wise performance on ID (test) slightly exceeded that of the OOD2 and was marginally inferior to OOD1, with a maximum difference of 2.9% in the $F1_w$ between ID vs OOD1. These findings were consistent for on-subject metrics. Noteworthy, on the ID test split, which contains "tastes" of more than 50 different physicians involved in scoring of PSGs, U-Sleep reached the subject-wise agreement level of κ = 76.2% that corresponds to the interscorer agreement of K = 76% reported in the literature.[12-14] This points to the robustness of U-Sleep's scoring ability in line with the theoretically justifiable performance ceiling that can be achieved on human-scored hypnograms. Marginal over- and under-performance on OOD data splits can be attributed to the greater or lesser consistency of the given split-specific senior physician with the "overall" population scoring pattern encoded in U-Sleep.

## *Evaluation of approaches for uncertainty estimation*

The primary objective in this phase was to pinpoint the best approach that adeptly identifies U-Sleep-predicted epochs that deviate from human scoring. This consisted of two main strands of investigation: comparing softmax-based uncertainty metrics and evaluating the confidence scores based on the adapted confidence neural network.

### Softmax-based measures

We initially took into consideration all the softmax-based metrics, as delineated in Table 3. The metrics (i-v) identify uncertain epochs based on a distributional threshold, while metrics (vi-vii) are designed to accumulate a predetermined percentage of the most uncertain predictions. The fixed-percentage strategies do not include an approach based on the softmax ratio ($\rho$) as it is monotonically dependent on the maximum of the softmax ($\mu$) and would lead to the same results. Calculation of these metrics was straightforward, as they involved only the U-Sleep softmax output based on each input channel-pair. Performance of individual measures in terms of identifying predictions discordant from human scoring is listed in Table 5. Majority of the metrics achieved comparable results with the superiority of the distributional-threshold-based metrics over the fixed-percentage strategies, confirming the need of a flexible approach adapting to possibly different amounts of difficult-to-score (uncertain) epochs per PSG. The best performing approach was $\mu$ - the maximum of the majority-softmax (= softmax averaged over all input channel pairs) - reaching AUROC of 76.5% on the ID-test and 82.4-81.1-% on the two OOD sets.

## Auxiliary Confidence Network

Our evaluations continued with the auxiliary confidence network leveraging the joint information of the transformed softmax output and the hidden representations extracted from U-Sleep to predict the *True Class Probability* (TCP) score. We trained the confidence network on the ID training and validation splits, targeting the actual TCP scores calculated based on predictions of the already trained U-Sleep classifier. The training was based on minimizing the mean-absolute-error (MAE) loss, adopting mini-batches of U-Sleep-derived features for one PSG channel pair (EEG-EOG) at the time, adhering to the default configurations of the Adam optimizer in Tensorflow 2.6.0. The training process achieved convergence after 16 epochs, marking a validation MAE of 0.0827. This indicates the confidence network's capability to predict the TCP with an average error of 8.27% in probabilistic terms. It is worth noting that the training set incorporated epochs labeled as "*unknown*" by physicians, reflecting the inherent challenges in scoring such signals, often due to untouched electrodes yielding constant (zero) signal. These particular epochs were assigned a target TCP of 0, given that none of the softmax values would match the correct class (ie, sleep-stage).

Having the trained confidence network we evaluated how its predicted TCP-score performs to detect discordant epochs. Focusing on the last column of Table 5, we observe its superiority in comparison to all simpler softmax-based approaches across all test data subsets. It outperformed the other approaches in terms of both ROC as well as PR assessments, reaching AUROC of 85.7% on ID, 85.6-82.5% on the two OOD sets, and AUPR of 63.1% for ID and 52.3-50.7%, respectively. Furthermore, the robustness of the confidence network was confirmed, as it delivered comparable performance on both ID and OOD splits, highlighting its generalizability to potentially different scoring patterns introduced by different senior physicians. Given its demonstrated efficacy, the TCP confidence score was selected as the key metric for the following evaluations simulating physician's interventions, focusing on the review and eventual correction of the most uncertain predictions.

**Confidence-supplemented hypnogram**

Using the TCP as the most reliable uncertainty quantification measure, Figure 2 depicts the combined output of the U-Sleep-predicted hypnogram (in white) with the estimated confidence TCP-scores as a green-red color scale in the background. This dual output is a result of our final pipeline, depicted as a diagram in Figure 3, detailing the process of transforming original biosignals into a joint presentation of predicted sleep stages and their associated confidence levels. Such visual representation is designed to guide the physician in identifying specific segments of the PSG that deserve closer review. For demonstration, the actual physician's scoring on given PSG, referred to as *True*, is depicted in blue. A close examination reveals that segments with lower predicted TCP scores often (eg, 1:30-2:30 h of sleep) predominantly align with U-Sleep misclassifications. In contrast, regions with higher scores (eg, from 3:30 h onwards) mostly point to accurately scored epochs. It is important to note that since the estimated TCP scores are model-derived, occasional discrepancies can arise. For instance, around 1:45 h, a brief period marked with high confidence corresponds to discordant scoring. Even though this segment erroneously indicates high confidence, its neighborhood areas of low confidence might draw physician's attention for a review. Despite occasional inconsistencies, the results from Table 5 indicate that TCP-score has the best ability to identify discordant epochs.

## *Statistical tests of on-subject TCP scores with respect to clinical diagnosis*

Further, we investigated in-depth the discriminative power of the TCP-score to reveal discordant predictions. Firstly, to evaluate $H_{01}$, we calculated the on-subject difference between averaged TCP-scores of predictions that align and those that disagree with human scoring: $d_i = \overline{TCP}_{i,correct} - \overline{TCP}_{i,incorrect}$. Next, for the evaluation of $H_{02}$, the on-subject performance metrics ($Acc_i$, $F1_{w,i}$, $K_i$) and the overall average TCP score ($\overline{TCP}_i$), for each subject's predicted hypnogram were calculated. The $\overline{TCP}_i$ can be understood as an assessment of the confidence over the entire predicted hypnogram of a given subject. We employed a non-parametric bootstrap approach, with 5000 repetitions, for both hypotheses to compute confidence intervals (CIs). Having a database rich in sleep-disorder diagnoses enabled us to assess both hypotheses considering individual classes of diagnoses, as described in Table 2. To assess generalizability of our findings, we considered subjects from the ID-test and the two OOD test data with confirmed conclusive diagnoses. Since the subjects - except for healthy controls - suffer in many cases from several sleep disorders, we always included in a given class all who have at least one corresponding diagnosis. Both hypotheses were assessed on disorder classes of at least 10 subjects, separately on the ID test data, and - to achieve a larger sample size in each class - the pooled OOD data.

Table 6 gives an overview of bootstrapped 95% CIs and the medians related to $H_{01}$ for each diagnosis class considered. Based on the CIs obtained, $H_{01}$ can be rejected (p-value < 0.05 in all cases) and one can conclude that the difference between the mean-aggregated TCP-scores of aligning and discordant predictions significantly differs and is consistently greater than 0. All that across the entire diagnosis spectrum, on both ID as well as OOD test domains. The median differences ranged as 0.20-0.23 and 0.19-0.26, for ID and OOD, respectively, which affirms that the TCP-score was in terms of probability about 20% lower for the discordant predictions. This suggests that the confidence network and the resulting TCP score can efficiently guide physicians on hypnogram and respective PSG sections needing review and potential correction, regardless of a patient's diagnoses status.

Further, Table 7 relates to $H_{02}$ and details the bootstrapped 95% CIs for the correlation between the average on-patient TCP score and the classification performance metrics. Based on the CIs obtained, we conclude that for all diagnoses of both ID and OOD test data, the correlation with any performance metric was consistently significant (p-value < 0.05 in all cases) and positive. The TCP correlated - on average - the most with the accuracy with a range of 0.67-0.74 across individual diagnosis classes of ID test data, and of 0.58-0.81 for OOD data. These findings suggest that the aggregated TCP score can efficiently pinpoint subjects whose biosignals are challenging to classify and also those with high prediction performance.

## *Performance boost under physician's intervention*

In this final part of our evaluations, we aimed to quantify the potential improvement in sleep-scoring classification performance when the most uncertain predictions underwent physician's review. We simulated an intervention in which predictions with a TCP confidence score falling below a designated threshold, incremented in 0.01 steps across the [0,1] range, were set aside for human assessment. Within this set, predictions that did not align with the physician's assessment were subsequently adjusted to reflect the physician's scoring evaluation. Alongside observing the uplift in performance, we also monitored the amount of predictions subjected to review. This amount is indicative of the physician's time spent on re-scoring, prompting us to quantify the effort needed to reach specific performance benchmarks.

Figure 4 depicts the impact of the physician's review on the classification performance for the ID-test and the two OOD test data. The lower x-axis depicts the TCP-score threshold used to gather uncertain predictions, whereas the upper x-axis to the corresponding total % of the epochs re-scored (ie, the physician's effort). The % refers to the aggregate over all PSGs in a given data split, as from each PSG were extracted only epochs below a given threshold and so, the individual % differed. At TCP-threshold of 0, when no uncertain epochs are extracted, the performance as depicted on the vertical axis corresponds to the original epoch-wise performance as shown in Table 4. From Figure 4, we can observe a monotonic improvement in all the performance metrics with the increasing amount of epochs gathered for the review. Based on that we can identify, that to reach, eg, at least 90% in all the evaluation metrics, the rescoring effort of about 26% for ID-test, 19% for OOD1, and 27% for OOD2 is needed, respectively, whereas the corresponding TCP threshold lies consistently around 0.75.

Further, based on Figure 4, Table 8 summarizes the % of epochs needed to be reviewed to achieve the performance benchmarks of at least (80, 85, 90, 95)% for each evaluation metric, which we use for the comparison with other existing works in the Discussion. For example, to reach at least 90% in K, a physician's review of 25.6% of epochs is needed on the ID-test, and 18.8%-29.0% on the two OOD datasets.

Finally, Figure 5 compares the rescoring effort based on an appropriate TCP-threshold in comparison to the % of all the misclassified epochs detected (ie, the true positive rate) per individual test data splits. The diagonal depicts a "random strategy", where physician's review would be conducted without any prior guidance on uncertain epochs. We observe that independently of the data domain, less than 50% of epochs need to be reviewed in order to detect at least 90% of all misclassified epochs. Similarly, to detect more than 95% of all misclassified epochs, the review of less than 60% of all epochs is needed. At a hypothetical 20% error rate, the 50% review effort with a corresponding detection of 90% out of all the discordant predictions, leads to a boost of 18% resulting in a scoring performance of 98%, conforming with proposed clinical standards and being far beyond acceptable scoring error rates.[15] Since in our case is the error rate less than 20% for all domains (accuracy is always > 80%, as indicated in Table 4), the 50% review effort corresponds to obtaining almost perfectly aligned hypnograms with agreement above 98%.

## Discussion

Our study was motivated by a key clinical application in the field of sleep medicine, where physicians reach a consensus of about 76% when scoring PSG into sleep stages.[12-14] This level of agreement sets a technical limit on accuracy metrics attainable when training scoring algorithms on multiple domains (scorers / databases). Consequently, when incorporating a scoring algorithm into clinical practice, its predictions must be subjected to a rigorous review by a human expert. If this is not guided to the uncertain regions of the predicted hypnogram and the respective PSG biosignals, such review may require a similar time effort as manual scoring done from scratch. Motivated by these challenges we designed a pipeline where a state-of-the-art scoring algorithm U-Sleep is adopted in combination with an uncertainty estimation approach to guide the human review on the predicted hypnograms, with a primary focus on the quantification of the effort required to achieve certain performance benchmarks. In order to robustly estimate the generalizability of our approach, we took benefit of the rich clinical database (BSDB),[22] and evaluated our efforts on both in-domain (ID) and the two out-of-domain (OOD) test data subsets and considering individuals' conclusive sleep-disorder diagnoses.

As a sleep scoring classifier, we adopted the well-established U-Sleep which we trained on 13 open-access databases and fine-tuned on ID (training and validation) data of BSDB. Such trained classifier reached a robust performance of K = 76.2% for ID test data, counting recordings scored by > 50 different physicians, and K = (78.8, 73.8)% on the two OOD sets, each being scored by a unique senior physician with potentially different scoring taste than the "population" scoring knowledge encoded in U-Sleep, respectively.

Following that, we conducted an extensive investigation of different uncertainty estimation approaches and assessed their performance on both ID and OOD test datasets. Remarkably, our designed auxiliary confidence network specifically trained for PSG time-series data, working in tandem with the U-Sleep, emerged as the superior approach, adeptly identifying predictions discordant with human scoring across both ID (AUROC=85.7%) and the two OOD test data (AUROC of 85.6-82.5%). Identifying an approach that accurately pinpoints disagreeing predictions was a key prerequisite to enabling efficient review of predicted hypnograms by physicians.

Furthermore, our research extended into statistical examinations of the predicted uncertainty estimates, namely confidence scores based on our auxiliary network, leading to two pivotal conclusions: (i) the on-subject confidence scores were significantly different and lower for epochs discordant with human scoring, and (ii) the on-subject aggregated confidence scores significantly and positively correlated with the on-subject classification performance metrics. Both findings were consistent over the entire spectrum of sleep diagnoses present in both ID and OOD test data. These insights not only validate the efficacy of our approach for physician's review but also highlight its capacity to pinpoint sections of PSG biosignals that are inherently challenging to score, independently of the subject's diagnosis status.

As a pivotal component of our evaluations, we examined the extent to which guiding physicians in reviewing uncertain epochs could augment the efficacy of sleep staging. To attain a commendable classification performance of at least 90% in (K, Acc, $F1_w$) metrics, our approach necessitated physicians to examine under 25.6% for K, 16.5% for Acc, and 17.2% for $F1_w$ of the epochs on ID test data. For both OOD data, these figures were less than 29.0%, 19.0%, and 21.9%, respectively. These outpace the findings by Hong et al (2021),[19] where about 35% and 25% of epochs needed a review to achieve a similar 90% rate in (K, $F1_w$) on ID data primarily from sleep-disordered subjects. In the broader context, the review effort of our approach closely mirrors that of Phan et al (2022).[20] In their study on the Sleep-EDF dataset of *healthy* subjects, they reported a requirement to review 50% of epochs to identify 90% of all misclassified epochs. In our setup, with a dataset predominantly featuring sleep-disordered subjects, our efforts resonated closely, demanding a review of 45-50% of epochs, on both ID as well as OOD test data. Notably, the review of 50% of all the epochs leads in our case agreement of >98% for all ID and OOD test datasets. Furthermore, aiming for a more stringent identification of 95% of all misclassifications, our approach stands out, demanding a review of less than 60% of epochs on both ID and OOD test data - a subtle improvement over the 61.4% reported by Hong et al (2021).[19] In addition, our efforts are in line with the findings of most recent work of Rusanen et al (2023),[21] who identified about 90% of all misclassified cases by reviewing 50% of all epochs on consensus-hypnograms of DOD database of 81 subjects (56 OSA + 25 healthy), where each PSG was scored by multiple experts. In our case, the level of this performance was achieved on ID as well as on two OOD single-scorer datasets of a considerably larger size containing subjects from a full spectrum of sleep-disorders. We consider results on our OOD datasets as remarkable positive since the adaptation of the approach to the scoring taste of a single scorer is expected to be more difficult for algorithms (U-Sleep, confidence network) trained on data containing scorings of different physicians, as it represents a change of domain from multiple- to a single-scorer one. Adapting to the single-scorer's taste is closer to the current setup in clinical practice, where obtaining multiple-scorers' consensus is costly, and a single physician evaluates the PSG and makes the final clinical decisions. These results spotlight not only the efficacy of our approach and its robustness to OOD data with different diagnosis statuses but also underscore the potential to reduce the physicians' workload on manual sleep staging, which is paramount in practical scenarios.

Yet, our work is not without limitations. The field of uncertainty quantification for sleep staging is relatively new and it does not include well-established baselines that would also incorporate publicly available data covering the full spectrum of sleep disorders. The data in the BSDB are mostly observational, ie, subjects undergo sleep studies due to suspicion or symptoms, and so, the presence of different diagnoses is not randomized or balanced. The training of both classification and uncertainty-estimation algorithms was done without explicit control for gender, ethnicity, age, and clinical diagnosis, which may - together with non-randomized data - contribute to computational bias.

## Conclusion

The significant challenges in automatic sleep staging, such as noise-amounts in the labels, inter-scorer disagreement, and heterogeneity of PSG databases - reflecting the large inter-individual variability in sleep manifestation - underscore the complexities in achieving an AI model that could perfectly generalize to data from different domains. While automated sleep scoring algorithms trained in a supervised way have achieved excellent performances despite these hurdles, they are still bound by the limitations inherent to the quality of their training labels. Consequently, despite the technological advancements, the critical role of physicians in reviewing and verifying predicted hypnograms remains - so far - irreplaceable and imperative. With the increasing prevalence of sleep-wake disorders, and with the massive amounts of data present in PSGs, it is necessary to drive research efforts to optimize physician's review by directing them to potential areas of uncertainty, while ensuring an efficient examination compliant with clinical needs.

In our study, we developed a pipeline aimed at enhancing the use of automated sleep-scoring algorithms in clinical practice. The retraining of the U-Sleep algorithm on almost 20000 PSGs encoded the scoring expertise of a broad range of physicians coming from 13 open-access databases. Utilizing the comprehensive BSDB database containing almost 9000 additional PSGs supplemented with clinical annotations, we compared various approaches for uncertainty quantification, including training of an auxiliary confidence network, which demonstrated its superiority for identifying predictions discordant from physician's scoring.

By uniquely combining the robust scoring capability of U-Sleep, the precision of added confidence network, and the richness of the BSDB database allowing validations with respect to individuals' conclusive diagnosis status, our study pave the way to successfully integrate the semi-automated approach into clinical practice. We also demonstrated the positive impact of the physician's review and quantified the effort required to achieve specific performance benchmarks across different data domains. The combined approach of our pipeline ensures that while insights from automatic scoring are utilized, physicians can concentrate their efforts on reviewing segments of biosignals where potential disagreements or algorithmic errors may occur. We believe that the adoption of scoring algorithms for clinical practice does not consist in replacing the physician's expertise with an algorithm, but mainly in enabling the effective use of the algorithm's insights and their thorough validation.

## Acknowledgements

The secondary usage of Berner Sleep Data Base (BSDB) from Inselspital, University Hospital Bern, was ethically approved (KER-Nr. 2020-01094) in the framework of the E12034 - SPAS (Sleep Physician Assistant System) Eurostar-Horizon 2020 program. The BSDB dataset access may be granted upon individual request, after data transfer agreements were put in place.

## Disclosure

The authors report no conflicts of interest in this work.

**Table 1** Demographic characteristics of BSDB with respect to individual data splits

| Domain | Scorers involved | Split | Number of PSGs | Age ($\mu \pm \sigma$) | Gender (% of males) |
|---|---|---|---|---|---|
| ID | > 8 SP, > 50 AP | train | 4245 | 49.22 ± 16.40 | 64.28 |
| | | validation | 226 | 52.66 ± 21.45 | 67.71 |
| | | test | 423 | 50.48 ± 20.32 | 65.57 |
| OOD1 | 1 SP | test | 1966 | 48.90 ± 18.60 | 64.65 |
| OOD2 | 1 SP | test | 1972 | 46.93 ± 20.06 | 60.92 |
| TOTAL | > 10 SP, > 50 AP | - | 8832 | 48.82 ± 18.25 | 63.76 |

**Abbreviations:** PSG, polysomnography; $\mu$, average age per group; $\sigma$, standard deviation of the age per group; ID, in-domain; OOD, out-of-domain. SP, senior physician; AP, assistant physician

**Table 2** Occurrence of different classes of sleep disorders among conclusive diagnoses of subjects per individual data splits of BSDB

| | Domain | | | | | |
|---|---|---|---|---|---|---|
| Diagnosis class | ID train | ID validation | ID test | OOD1 test | OOD2 test | ALL |
| **HE** | 27 | 2 | 3 | 12 | 22 | 66 |
| **INS** | 106 + 15 | 8 + 1 | 17 + 2 | 31 + 5 | 43 + 4 | 205 + 27 |
| **SDB** | 247 + 156 | 16 + 8 | 34 + 17 | 91 + 33 | 124 + 18 | 512 + 232 |
| **CDH** | 171 + 30 | 10 + 2 | 22 + 5 | 54 + 1 | 115 + 10 | 372 + 48 |
| **CRD** | 11 + 1 | 0 + 0 | 2 + 0 | 1 + 0 | 5 + 0 | 19 + 1 |
| **PSD** | 75 + 9 | 7 + 0 | 6 + 0 | 22 + 1 | 44 + 0 | 154 + 10 |
| **SMD** | 74 + 5 | 5 + 0 | 7 + 0 | 18 + 1 | 33 + 0 | 137 + 6 |
| **IS** | 227 + 11 | 13 + 1 | 26 + 1 | 77 + 1 | 127 + 1 | 470 + 15 |
| **DSS** | 26 + 0 | 1 + 0 | 2 + 0 | 7 + 0 | 16 + 0 | 52 + 0 |
| **Multiple disorders** | 418 | 26 | 52 | 128 | 205 | 829 |
| **Single disorders** | 227 | 12 | 25 | 42 | 33 | 339 |
| **Other or unknown** | 3573 | 186 | 343 | 1784 | 1712 | 7598 |
| **TOTAL** | 4245 | 226 | 423 | 1966 | 1972 | 8832 |

**Notes:** Columns indicate individual data subsets: ID (training, validation, testing) and two OOD test sets (OOD1, OOD2), summing up to ALL. Rows indicate the number of subjects according to conclusive diagnoses class indicated by abbreviations described below. Row **Multiple disorders** indicates the number of subjects with multiple classes of sleep-disorders, **Single disorder** the number of subjects with a single sleep disorder, and **Other or unknown** the number of subjects with no or unknown conclusive diagnosis. **TOTAL** is equal to HE + Multiple disorders + Single disorder + Other or unknown. At the cell level of rows (INS to DSS), the sum refers to the number of subjects having multiple disorders including that given class plus the number of subjects having that specific class only.

**Abbreviations:** ID, in-domain; OOD, out-of-domain; HE, healthy controls; INS, insomnia disorders; SDB, sleep-disordered breathing; CDH, central disorders of hypersomnolence; CRD, circadian rhythm sleep-wake disorders; PSD, parasomnia-related sleep disorders; SMD, sleep-related rhythmic movement disorders; IS, isolated symptoms and normal variants; DSS, findings specific to day-time sleep studies.

**Table 3** Measures evaluating prediction's uncertainty using U-Sleep softmax output

| ID | Measure | Notation | Mathematical formula |
|---|---|---|---|
| i | Average softmax-entropy | $\overline{p}_{entr}$ | $-\frac{1}{M}\sum_{m=1}^{M}\sum_{k=1}^{5}p_{mk}\log_2 p_{mk}$ |
| ii | Average softmax-ratio | $\overline{\rho}$ | $\frac{1}{M}\sum_{m=1}^{M}\frac{1}{5}\sum_{k=1}^{5}\frac{p_{mk}}{max(p_m)}$ |
| iii | Average softmax-standard-deviation | $\overline{\sigma}$ | $\frac{1}{M}\sum_{m=1}^{M}\frac{1}{4}\sum_{k=1}^{5}\sqrt{(p_{mk}-\frac{\sum_{k=1}^{5}p_{mk}}{5})^2}$ |
| iv | Maximum of majority-softmax | $\mu$ | $max(p_M) = max(\frac{1}{M}\sum_{m=1}^{M}p_m)$ |
| v | Standard deviation of majority-softmax | $\sigma$ | $\frac{1}{4}\sum_{k=1}^{5}\sqrt{(p_{Mk}-\frac{\sum_{k=1}^{5}p_{Mk}}{5})^2}$ |
| vi | Fixed % according to $\mu$ | $\mu$% | - |
| vii | Fixed % according to $\sigma$ | $\sigma$% | - |

**Notes:** Uncertainty measures adapted for majority-voting mechanism of the U-Sleep classifier.

**Abbreviations**: $M$ = total number of EEG-EOG channel-pairs used, $m$ = index over $M$, $k$ = index over 5 classes (sleep stages), $p_{mk}$ = probability (softmax value) of the $k$-th class based on the $m$-th input channel pair.

**Table 4** Classification performance of U-Sleep on individual data splits

| Domain | Metric | Epoch-wise | Subject-wise mean | Subject-wise median |
|---|---|---|---|---|
| **ID-test** | Acc | 82.5 | 82.1 | 84.5 |
| | F1$_w$ | 82.8 | 82.4 | 85.3 |
| | K | 75.0 | 71.2 | 76.2 |
| **OOD1** | Acc | 84.2 | 84.5 | 86.4 |
| | F1$_w$ | 85.0 | 85.5 | 87.4 |
| | K | 77.6 | 76.0 | 78.8 |

| Domain | Metric | Epoch-wise | Subject-wise mean | Subject-wise median |
|---|---|---|---|---|
| **ID-test** | Acc | 82.5 | 82.1 | 84.5 |
| | F1$_w$ | 82.8 | 82.4 | 85.3 |
| | K | 75.0 | 71.2 | 76.2 |
| **OOD2** | Acc | 80.7 | 80.8 | 82.7 |
| | F1$_w$ | 80.5 | 81.4 | 83.4 |
| | K | 73.3 | 71.1 | 73.8 |

**Notes:** Epoch-wise performance calculated over all 30-second windows present in individual data splits. Mean and median subject-wise metrics are calculated as performance achieved on individual-specific hypnograms.

**Abbreviations:** ID, in-domain; OOD, out-of-domain; Acc, accuracy; F1$_w$, weighted F1-score; K, Cohen's kappa.

**Table 5** Performance of uncertainty measures to identify U-Sleep predictions discerning from human scoring on individual data splits.

| Domain | Evaluation metric | Uncertainty measure | | | | | | | |
|---|---|---|---|---|---|---|---|---|---|
| | | $\bar{p}_{entr}$ | $\bar{\rho}$ | $\bar{\sigma}$ | $\mu$ | $\sigma$ | $\mu\%$ | $\sigma\%$ | TCP* |
| **ID-test** | AUROC | 76.4 | 75.7 | 76.2 | 76.5 | 64.3 | 59.1 | 56.5 | **85.7*** |
| | AUPR | 39.7 | 41.3 | 41.0 | 42.9 | 30.2 | 36.5 | 31.4 | **63.1*** |
| **OOD1** | AUROC | 80.1 | 82.0 | 81.6 | 82.4 | 75.4 | 60.6 | 57.2 | **85.6*** |
| | AUPR | 38.8 | 42.0 | 41.0 | 43.5 | 41.0 | 33.3 | 26.8 | **53.6*** |
| **OOD2** | AUROC | 79.6 | 80.8 | 80.6 | 81.1 | 75.0 | 59.9 | 57.1 | **82.5*** |
| | AUPR | 43.2 | 45.0 | 44.6 | 45.8 | 34.1 | 36.9 | 31.4 | **50.7*** |

**Notes:** Performance assessment as the % of the AUROC and AUPR curves for the softmax-based measures from Table 3 and the *True Class Probability* (TCP) score based on confidence network.

**Abbreviations:** ID, in-domain; OOD, out-of-domain; AUROC, area under receiver operating characteristic (curve); AUPR, area under precision-recall (curve); TCP, true class probability; uncertainty measures consistent with Table 3.

**Table 6** Bootstrap confidence intervals for difference of on-subject mean-aggregated confidence TCP-scores of aligning vs discordant predictions

|  | ID-test | | | Pooled OOD | | |
| --- | --- | --- | --- | --- | --- | --- |
| Diagnosis class | Median | 95% CI | N | Median | 95% CI | N |
| HE |  |  | NA | 0.26 | (0.22, 0.30) | 34 |
| INS | 0.21 | (0.18, 0.24) | 19 | 0.23 | (0.21, 0.25) | 83 |
| SDB | 0.20 | (0.17, 0.22) | 51 | 0.21 | (0.20, 0.22) | 266 |
| CDH | 0.23 | (0.20, 0.27) | 27 | 0.23 | (0.21, 0.24) | 180 |
| PSD |  |  | NA | 0.19 | (0.16, 0.21) | 67 |
| SMD |  |  | NA | 0.20 | (0.18, 0.23) | 52 |
| IS | 0.21 | (0.18, 0.25) | 27 | 0.22 | (0.21, 0.23) | 206 |
| DSS |  |  | NA | 0.25 | (0.21, 0.29) | 23 |

**Notes:** Evaluations on ID-test data and pooled OOD data. Median stands for the estimate of the mean-difference and corresponding 95% CI are calculated as 2.5% and 97.5% quantiles of bootstrap resamples.

**Abbreviations:** ID, in-domain; OOD, out-of-domain; CI, confidence interval; N, number of subjects; NA, not available;; diagnosis classes consistent with Table 2.

**Table 7** Bootstrap confidence intervals for correlation between on-subject mean-aggregated confidence TCP-scores and the performance metrics

|  |  | ID-test | | | Pooled OOD | | |
| --- | --- | --- | --- | --- | --- | --- |
| Diagnosis class | Performance metric | Median | 95% CI | N | Median | 95% CI | N |
| HE | Acc<br>K<br>F1$_w$ |  |  | NA | 0.74<br>0.67<br>0.60 | (0.59, 0.91)<br>(0.50, 0.89)<br>(0.41, 0.84) | 34 |
| INS | Acc | 0.67 | (0.46, 0.91) | 19 | 0.58 | (0.46, 0.78) | 83 |
| | K<br>F1$_w$ | 0.56<br>0.63 | (0.28, 0.88)<br>(0.39, 0.91) | | 0.59<br>0.49 | (0.47, 0.78)<br>(0.36, 0.72) | |
| **SDB** | Acc<br>K<br>F1$_w$ | 0.71<br>0.69<br>0.57 | (0.62, 0.85)<br>(0.58, 0.85)<br>(0.44, 0.78) | 51 | 0.71<br>0.68<br>0.65 | (0.66, 0.80)<br>(0.62, 0.77)<br>(0.59, 0.75) | 266 |
| **CDH** | Acc<br>K<br>F1$_w$ | 0.72<br>0.64<br>0.58 | (0.55, 0.90)<br>(0.45, 0.86)<br>(0.35, 0.85) | 27 | 0.75<br>0.72<br>0.68 | (0.69, 0.84)<br>(0.66, 0.82)<br>(0.61, 0.79) | 180 |
| **PSD** | Acc<br>K<br>F1$_w$ | | | NA | 0.81<br>0.81<br>0.78 | (0.74, 0.89)<br>(0.74, 0.90)<br>(0.70, 0.87) | 67 |
| **SMD** | Acc<br>K<br>F1$_w$ | | | NA | 0.63<br>0.54<br>0.55 | (0.47, 0.84)<br>(0.37, 0.79)<br>(0.39, 0.79) | 52 |
| **IS** | Acc<br>K<br>F1$_w$ | 0.74<br>0.74<br>0.62 | (0.58, 0.90)<br>(0.59, 0.90)<br>(0.40, 0.87) | 27 | 0.70<br>0.64<br>0.63 | (0.64, 0.80)<br>(0.57, 0.76)<br>(0.56, 0.75) | 206 |
| **DSS** | Acc<br>K<br>F1$_w$ | | | NA | 0.62<br>0.62<br>0.49 | (0.34, 0.92)<br>(0.36, 0.93)<br>(0.21, 0.86) | 23 |

**Notes:** Evaluations on ID-test data and pooled OOD data. Median stands for the estimate of correlation with a performance metric and corresponding 95% CI are calculated as 2.5% and 97.5% quantiles of bootstrap resamples.

**Abbreviations:** ID, in-domain; OOD, out-of-domain; CI, confidence interval; N, number of subjects; NA, not available; K, Cohen's kappa; Acc, accuracy; F1$_w$, weighted F1-score; diagnosis classes consistent with Table 2.

**Table 8** Rescoring amounts needed to achieve desired levels of sleep-scoring performance

| | | Desired scoring performance level | | | |
|---|---|---|---|---|---|
| **Domain** | **Metric** | **80%** | **85%** | **90%** | **95%** |
| **ID-test** | K | 7.6 | 15.7 | 25.6 | 41.5 |
| | Acc | 0 | 6.0 | 16.5 | 32.2 |
| | F1$_w$ | 0 | 6.2 | 17.2 | 33.7 |
| **OOD1** | K | 2.3 | 9.1 | 18.8 | 32.9 |
| | Acc | 0 | 1.1 | 9.8 | 25.5 |

|  | F1$_w$ | 0 | 0.1 | 10.5 | 25.5 |
|---|---|---|---|---|---|
| **OOD2** | K | 8.3 | 17.3 | 29.0 | 44.0 |
|  | Acc | 0 | 6.7 | 19.0 | 37.0 |
|  | F1$_w$ | 0 | 8.3 | 21.9 | 39.1 |

**Abbreviations:** ID, in-domain; OOD, out-of-domain; K, Cohen's kappa; Acc, accuracy; F1$_w$, weighted F1-score

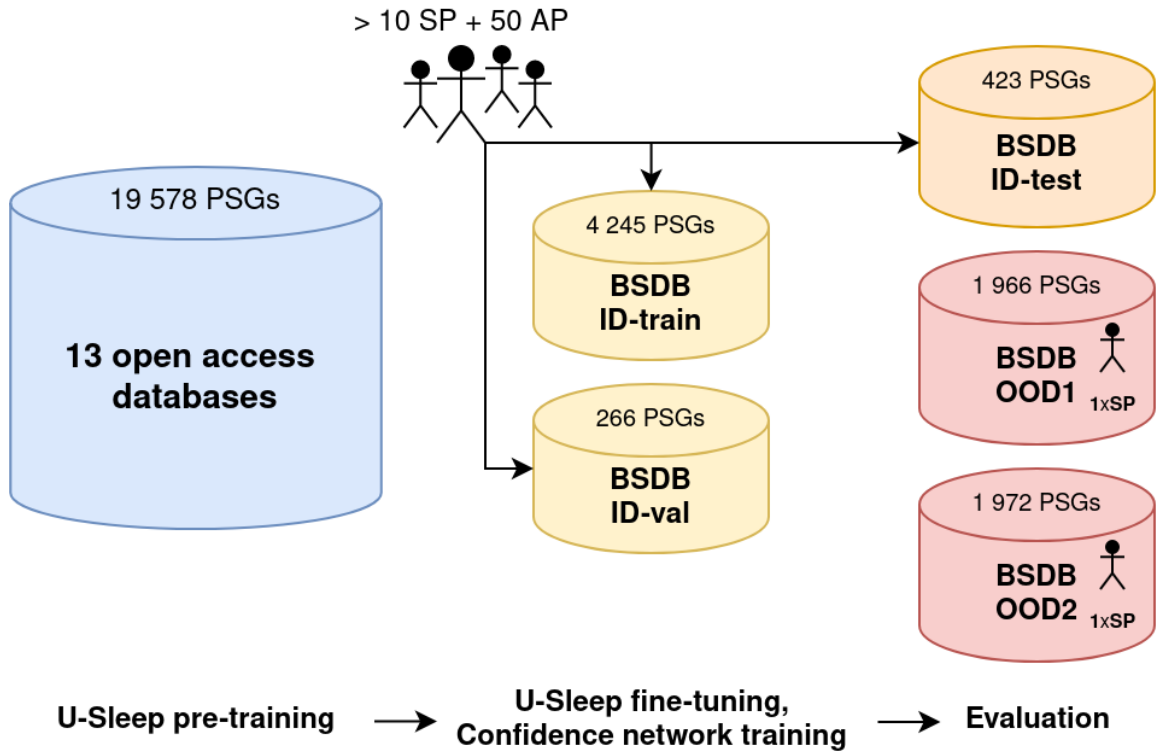

**Figure 1:** Schematic overview of datasets used, their size, and purpose.

**Notes:** The set of 13 open-access datasets (in blue on the left) was used for the baseline training of the U-Sleep. The middle and right parts of the schema relate to the evaluations on BSDB. Its ID part refers to PSGs each scored by one of more than 50 assistant and 10 senior physicians. The ID training and validation splits (in yellow) were used to fine-tune U-Sleep and, subsequently, to train the confidence network. Baseline evaluation of both algorithmic approaches was performed on the ID-test data (in orange). Their robustness was further evaluated on two OOD test sets (in red), each containing PSGs scored by a unique SP.

**Abbreviations:** ID, in-domain; OOD, out-of-domain; SP, senior physician; AP, assistant physician; PSG, polysomnography.

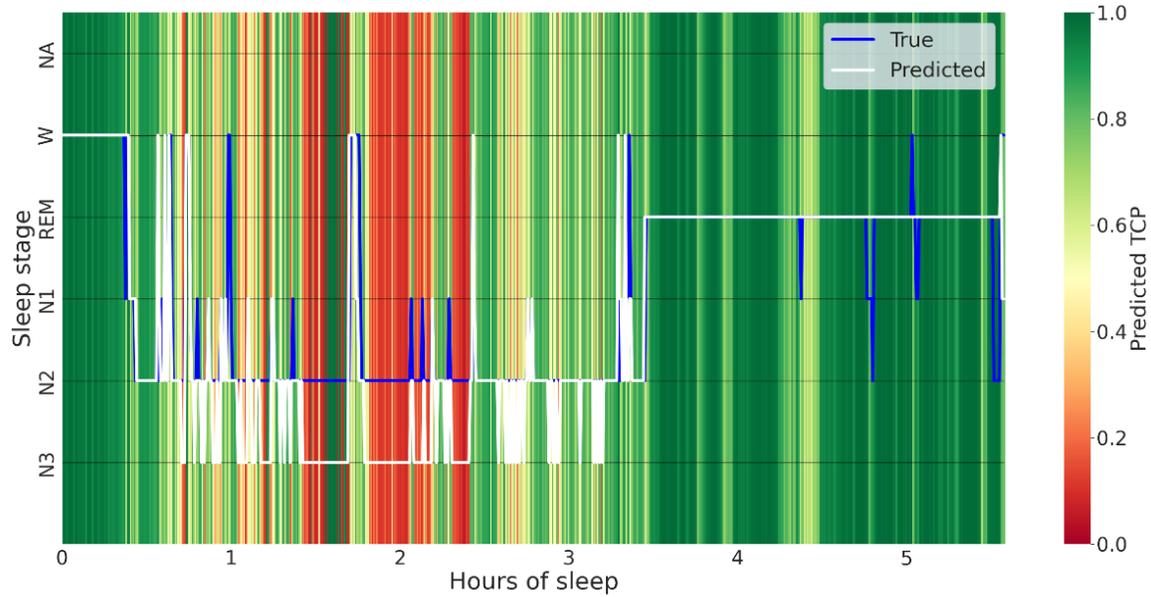

**Figure 2:** Combined output of the predicted hypnogram (in white) and the associated confidence.

TCP-scores (in the background), supplemented with the physician-scored hypnogram (in blue).

**Notes:** Combined output for a 44 years old female diagnosed with hypersomnolence. On-subject (Acc, $F1_w$, K) of (79.2, 72.2, 61.5)%, respectively. On-subject average TCP of 0.74. For correctly and incorrectly classified epochs, the average on-subject TCP was 0.87 and 0.41, respectively.

**Abbreviations:** TCP, true class probability.

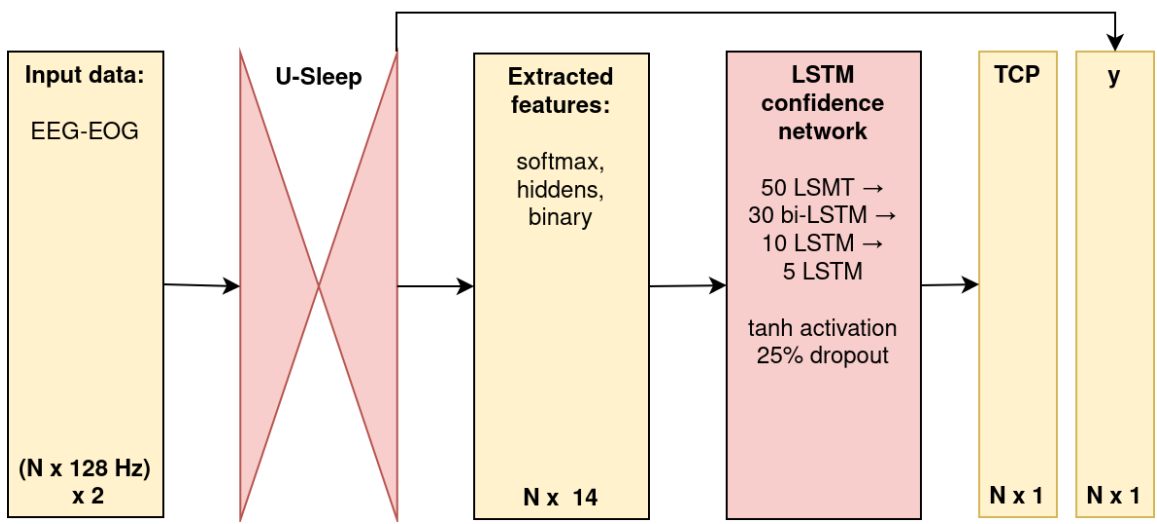

**Figure 3:** Schematic overview of the implemented pipeline.

**Notes:** An EEG-EOG channel-pair is used as an input for the U-Sleep classifier. Using the trained U-Sleep, several representations are extracted (*softmax*; *binary* code indexing the predicted class; hidden representations - *hiddens* - from the layer preceding softmax) and used as an input for the confidence network evaluating the True Class Probability (TCP) confidence score. The hypnogram predicted by U-Sleep (y) is provided jointly with the assessment of predictive uncertainty (1-TCP) to guide an efficient review by physician.

**Abbreviations:** N, number of epochs; TCP, true class probability; y, U-Sleep predicted sleep-stages.

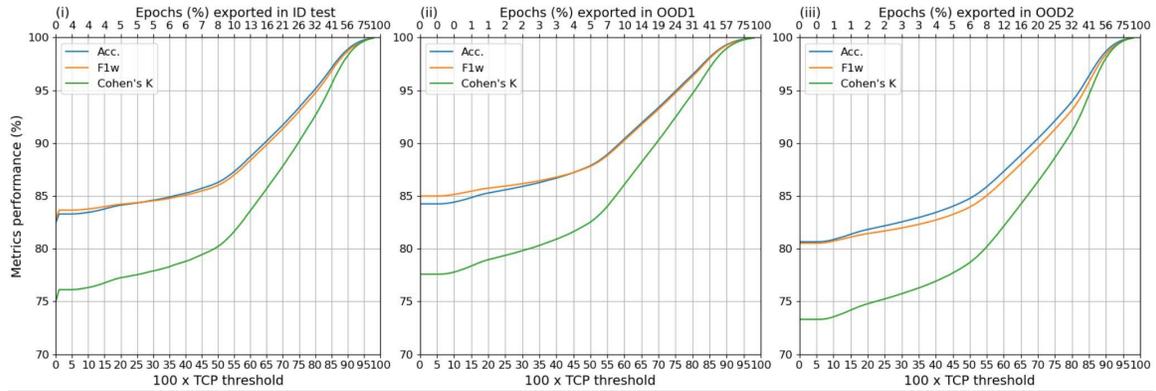

**Figure 4:** Performance boost with physician's review of epochs having confidence TCP-score lower than a given threshold.

**Abbreviations:** ID, in-domain; OOD, out-of-domain; K, Cohen's kappa; Acc, accuracy; F1$_w$, weighted F1-score.

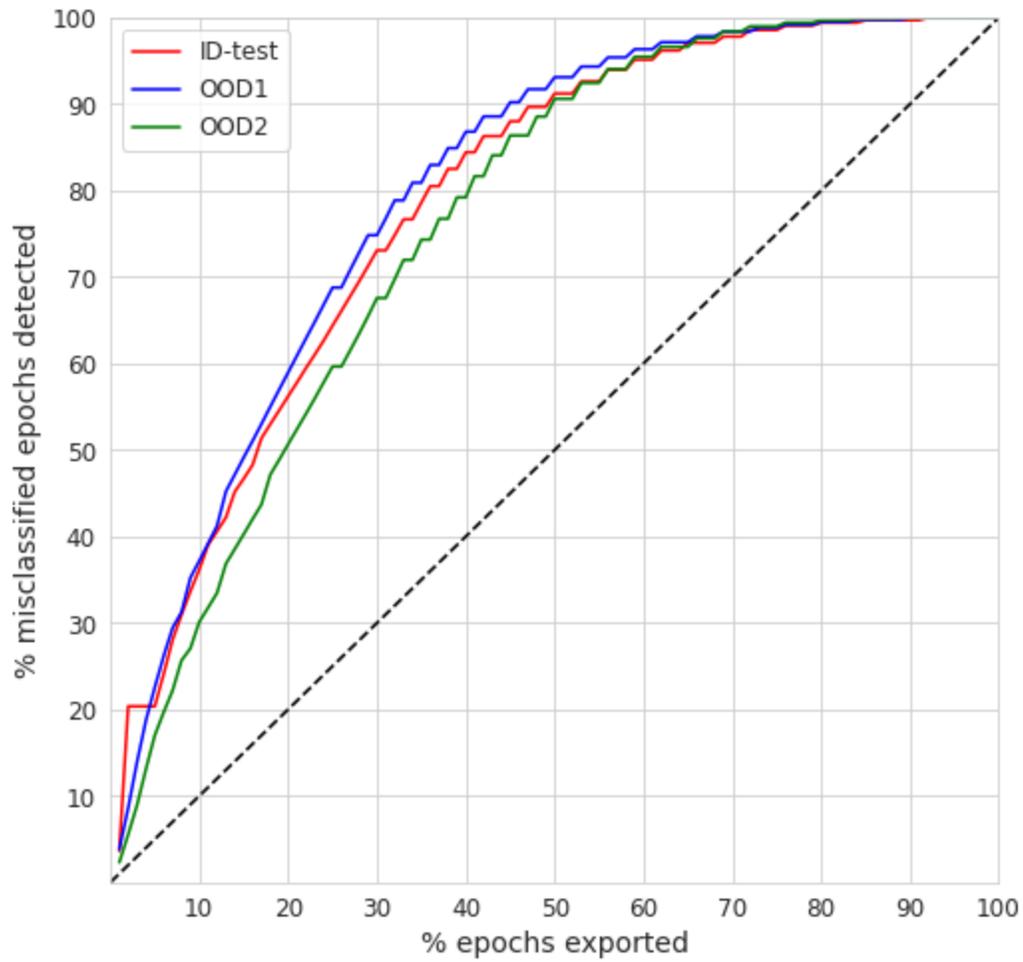

**Figure 5:** Review amounts (% of epochs exported) versus the % of all discordant predictions gathered.

**Abbreviations:** ID, in-domain; OOD, out-of-domain.